# A Hybrid Model for Traffic Incident Detection based on Generative Adversarial Networks and Transformer Model

**Xinying Lu**
School of Optical-Electrical and Computer Engineering
University of Shanghai for Science and Technology, Shanghai, China, 200093
Email: 222240474@st.usst.edu.cn

**Doudou Zhang**
Principal Data Scientist
Capital One
Mclean, VA, USA, 22102
Email: doudouzhang181@gmail.com

**Jianli Xiao (Corresponding Author)**
School of Optical-Electrical and Computer Engineering
University of Shanghai for Science and Technology, Shanghai, China, 200093
Email: audyxiao@sjtu.edu.cn

Word Count: 5300 words + 7 table (250 words per table) = 7050 words

*Submitted [Submission Date]*

**ABSTRACT:**

In addition to enhancing traffic safety and facilitating prompt emergency response, traffic incident detection plays an indispensable role in intelligent transportation systems by providing real-time traffic status information. This enables the realization of intelligent traffic control and management. Previous research has identified that apart from employing advanced algorithmic models, the effectiveness of detection is also significantly influenced by challenges related to acquiring large datasets and addressing dataset imbalances. A hybrid model combining transformer and generative adversarial networks (GANs) is proposed to address these challenges. Experiments are conducted on four real datasets to validate the superiority of the transformer in traffic incident detection. Additionally, GANs are utilized to expand the dataset and achieve a balanced ratio of 1:4, 2:3, and 1:1. The proposed model is evaluated against the baseline model. The results demonstrate that the proposed model enhances the dataset size, balances the dataset, and improves the performance of traffic incident detection in various aspects.

**Keywords**：Traffic incident detection, Generative adversarial networks, Transformer, Traffic incident dataset

**INTRODUCTION**

The expansion of industrial scale and the promotion of urbanization have become prominent features of nations worldwide. Expressways have become the preferred mode of transportation for short and medium distances due to their flexibility, speed, and efficiency. Expressways are also associated with a higher incidence of traffic accidents compared to other transportation modes. Freeway traffic incidents encompass various incidents that result in a reduction in road capacity or an abnormal surge in traffic demand. Traffic incidents refer to non-periodic incidents in which the location and timing are mostly unpredictable. They primarily consist of abnormal incidents that disrupt traffic flow and cause congestion, such as traffic accidents, vehicle breakdowns, road repairs, and extreme weather conditions. Automated detection of traffic incidents is one of the core functionalities of intelligent transportation systems. The automatic detection of traffic incidents is crucial for ensuring the smooth operation of urban traffic and improving the safety of people's travel. The challenge lies in rapidly and accurately detecting traffic incidents while avoiding false alarms. To address these challenges, various algorithms for traffic incident detection have been proposed.

In 1965, the California algorithm (*1*) was developed to detect traffic incidents based on the increasing and decreasing share of upstream and downstream traffic incidents. In 1970, the standard deviation algorithm (*2*) used the standard deviation value to judge the changing trend of traffic flow parameters and the observed average trend in previous time periods and judged the occurrence of traffic incidents if it was too large. In 1978, the Bayesian algorithm introduced probability to represent traffic time (*3*). In 1990, McMaster's algorithm based on mutation theory added traffic congestion into the object of judgment. After 1990, various artificial intelligence algorithms began to be applied in traffic incident detection, such as artificial neural network (ANN) (*4*), probabilistic neural network (PNN) (*5*), multi-layer perceptron (MLP) (*6*), fuzzy neural network (FNN) (*7*) and achieved good detection results in traffic incident detection. In recent years, deep learning models such as convolutional neural networks (CNNs) (*8*), adversarial generative networks (GANs) (*9*), and graph convolutional neural networks (GCNs) (*10*) have been applied to traffic incident detection to improve detection performance. Advancements in science and technology have facilitated the analysis of traffic flow parameters and spurred the development of cutting-edge technologies for traffic incident detection.

In the field of traffic incident detection, various approaches have been explored, ranging from probabilistic algorithms to neural network algorithms, with more recent applications of deep learning models. For a traffic incident detection system, precise detection of incidents, effective reduction of false alarms and missed detections, and achieving faster real-time detection are of paramount importance to enhance the reliability, stability, and practical application of the system. The application of advanced algorithms can effectively improve traffic management and enhance the efficiency and sustainability of urban transportation. Previous research has found that, in addition to utilizing more advanced algorithms, data acquisition, dataset size, and dataset balance also significantly impact the performance of detection models. Dataset imbalance can lead to a bias towards non-incident samples, resulting in reduced sensitivity to incident samples. Therefore, the key research challenge lies not only in applying advanced algorithms but also in obtaining large and balanced datasets.

In this paper, a hybrid approach combining GANs and a transformer model was proposed for traffic incident detection. Specifically, the GANs model was initially employed to address the challenges of limited sample size and dataset imbalance. Furthermore, the transformer model was integrated with the GANs model to overcome these challenges. This approach aims to leverage the strengths of both models and provide a comprehensive solution to the research difficulties. By harnessing the advantages of GANs and transformers, this hybrid model aims to improve detection performance by effectively handling sample scarcity and imbalance. The GANs contribute to generating diverse and balanced samples, while the transformer model enables deep feature extraction and captures complex relationships within the data. This innovative approach holds promise in enhancing the accuracy and robustness of traffic incident detection systems.

**PROPOSED MODEL**

This section begins with an introduction to the GANs, employed as a solution to address the issue of sample imbalance. Following that, the fundamental principles of the transformer model are elaborated upon. Finally, a novel integrated model is proposed, which combines the aforementioned models, providing a comprehensive approach to address the research problem.

**Generative Adversarial Networks**

The GANs consist of two essential components, namely the generator and the discriminator. The primary objective of GANs is to generate synthetic data that closely resembles real data by simulating the underlying data distribution (*11*). The generator is responsible for generating synthetic samples that mimic the characteristics of the real data. It takes as input a random noise vector or a real sample and employs a deep neural network to transform it into a "fake" sample. The generator aims to learn the mapping function that transforms the noise vector into a sample that convincingly resembles the real data. Through an iterative training process, the generator becomes increasingly proficient at generating more realistic samples.

The discriminator acts as a classifier that discriminates between real and synthetic samples. It takes as input both real and fake samples and predicts whether the sample is genuine or artificially generated. The discriminator is trained to optimize its ability to differentiate between the two types of samples accurately (*12*). Simultaneously, the generator seeks to deceive the discriminator by producing synthetic samples that are indistinguishable from real ones. This adversarial game between the generator and the discriminator forms the crux of the GANs framework. The optimization objective of GANs can be formulated as follows:

$$\min_G \max_D (D,G) = E_{X \sim P_{data}}[\log D(x)] + E_{Z \sim P_Z(Z)}[\log(1-D(G(Z)))], \quad (1)$$

where $G$ represents the generator, $D$ denotes the discriminator, $x$ represents real data samples, $z$ represents random noise vectors, and $p_{data}$ and $p_z(z)$ represents the data distribution and noise distribution, respectively. The $log D(x)$ represents the discriminant apparatus capable of identifying real samples, and $D(G(z))$ represents the generator cheat discriminant ability.

In the training process, the generator and discriminator are trained alternately, first $D$, then $G$, over and over again. The optimization process of $G$ and $D$ is based on the following formula:

$$Loss_D = \frac{1}{m} \sum_{i=1}^{m} (\log D(x_i) + \log(1-D(z_i))), \quad (2)$$

In the training of the discriminator, the objective is for it to correctly classify real incident samples *x* as 1 and generated incident samples *G(z)* as 0. Therefore, we expect *D(x)* to approach 1 and *D(G(z))* to approach 0. By substituting these objectives into **Equation 2**, the optimal loss value becomes 0.

$$Loss_G = \frac{1}{m}\sum_{i=1}^{m}(1-D(G(z_i))). \tag{3}$$

In the training of the generator, the objective is to have the discriminator classify the generated samples *G(z)* as real incident samples. Therefore, we want *D(G(z))* to approach 1, indicating that the generated samples are indistinguishable from real incident samples. By substituting this objective into **Equation 3**, the optimal loss tends to negative infinity.

where *m* is the number of training samples. The objective function aims to simultaneously maximize the accuracy of the discriminator in correctly classifying real and synthetic samples and minimize the accuracy of the discriminator in distinguishing between them. Through this minimax game, the generator gradually learns to generate samples that are increasingly difficult for the discriminator to differentiate from real data, while the discriminator becomes more adept at discriminating between real and synthetic samples.

By optimizing the GANs model iteratively, the generator and discriminator engage in a competitive process that drives the overall model to generate synthetic samples that closely resemble the real data distribution. This enables GANs to produce highly realistic and novel data instances, making them a powerful tool for various applications such as image synthesis, text generation, and data augmentation, among others. The architecture of the GANs model is shown in **Figure 1**:

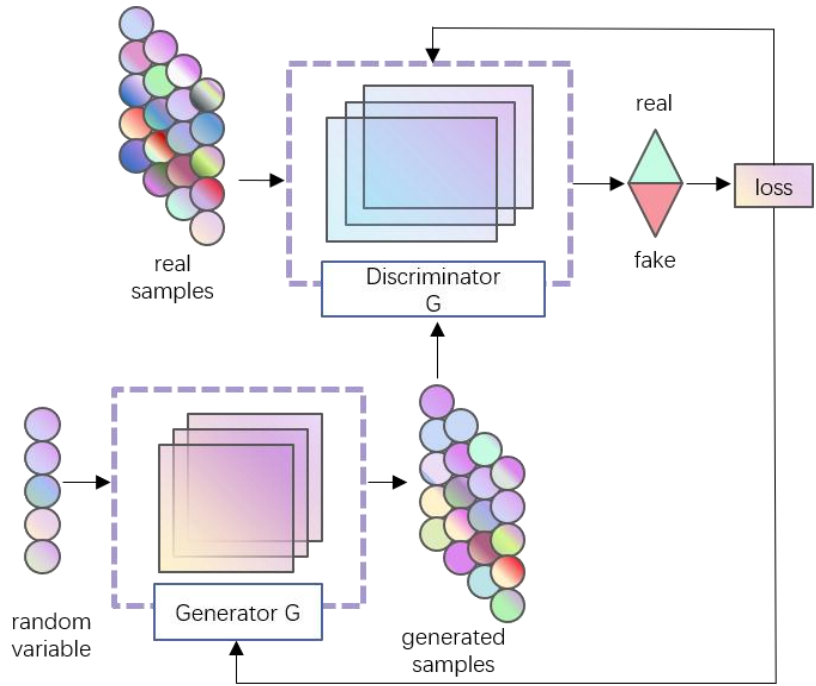

**Figure 1 The architecture of the GANs model**

**Transformer Model**

The transformer model has revolutionized natural language processing by introducing a groundbreaking architecture that captures long-range dependencies and enables parallel computation (*13*). Its key components are the encoder and decoder. The encoder comprises multiple identical layers, each consisting of a self-attention mechanism and feed-forward neural networks. The self-attention mechanism allows the model to establish connections between different positions in the input sequence, capturing interactions among tokens and enabling a better understanding of the sequence (*14*). The architecture of the transformer model is shown in **Figure 2**:

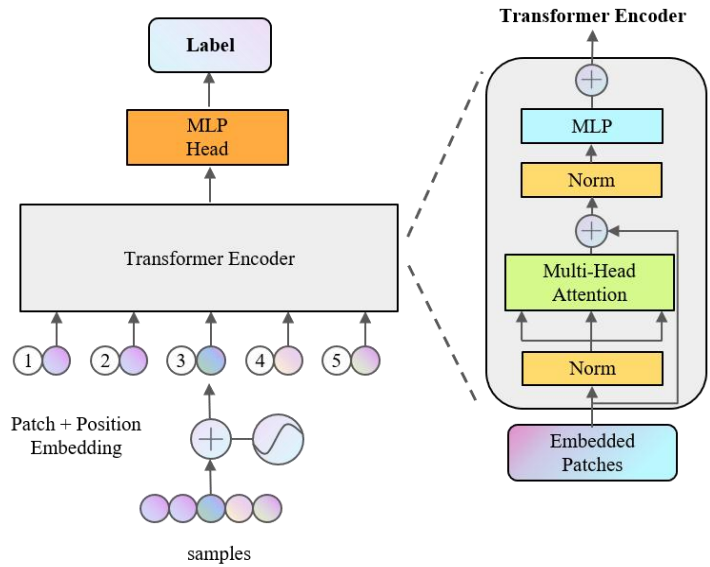

**Figure 2 The architecture of the Transformer model**

Positional encoding, another crucial component, models the positional information of input tokens. It introduces sequence order by assigning a positional vector to each token, enabling the model to differentiate between positions and capture their relative positions. As shown in **Figure 2**, each variable in the input sequence is assigned positional information, and positional encoding is used to represent the encoding vectors for each position in the sequence (*15*). These vectors are added to the input embedding vectors, allowing the model to differentiate between different positions of tokens and capture their relative positions within the sequence.

The specific form of positional encoding is calculated using the following formula:

$$PE_{(pos,2i)} = \sin\left(\frac{pos}{100^{\frac{2i}{d_{model}}}}\right), \qquad (4)$$

$$PE_{(pos,2i+1)} = \cos\left(\frac{pos}{100^{\frac{2i}{d_{model}}}}\right), \qquad (5)$$

here, $PE_{(pos,2i)}$ represents the positional encoding value at position pos and dimension $2i$, and $PE_{(pos,2i+1)}$ represents the positional encoding value at position pos and dimension $2i+1$. The term $d_{model}$ represents the dimension of the embedding vector, and $i$ represents the dimension index in the positional encoding. The sine and cosine functions in the formula introduce periodicity in the positional encoding along the dimensions. The term $1000^{\frac{2i}{d_{model}}}$ controls the relative scaling factor for different dimensions, resulting in different dimensions of positional encoding having different periodicity.

A notable advantage of the transformer model is its ability to handle long-range dependencies in classification tasks (*16*). Transformer employ self-attention mechanisms that enable tokens to attend to any other token in the sequence. This capability empowers the model to capture distant relationships and contextual dependencies, leading to improved classification performance. The self-attention mechanism is a key component of the transformer model, allowing it to capture the relationships between different positions within an input sequence (*17*). It achieves this by mapping the input sequence into multiple subspaces and performing attention computations within each subspace, allowing the model to learn and capture different relationships in parallel, as shown in **Figure 3**.

The input sequence is linearly transformed to obtain representations for queries (*Q*), keys (*K*), and values (*V*). They are calculated using the following formula:

$$\begin{aligned}
Q &= XW_Q \\
K &= XW_K \qquad (6) \\
V &= XW_V,
\end{aligned}$$

here, $X$ represents the input sequence, and $W_Q$, $W_K$, and $W_V$ are learned weight matrices. Next, the $Q$, $K$, and $V$ are split into h heads, as the following formula:

$$\begin{aligned}
Q_j &= QW_{\{Q_j\}} \\
K_j &= KW_{\{K_j\}} \qquad (7) \\
V_j &= VW_{\{V_j\}},
\end{aligned}$$

here, $j$ denotes the *j-th* head, and $W_{\{Q_j\}}$, $W_{\{K_j\}}$, and $W_{\{V_j\}}$ are the weight matrices specific to each head. Attention weights are computed for each head, calculating as the following formula:

$$Attweight_j = soft\max\left(\frac{Q_j K_j^T}{\sqrt{d_K}}\right), \qquad (8)$$

the value $d_k$ represents the dimension of each attention head, which is typically set to the same value as the dimension of the input embeddings divided by the number of attention heads. In the attention calculation, the query matrix $Q$ is used to compute the similarity between the positions in the sequence. The key matrix $K$ is used to represent the importance of each position, and the value matrix $V$ contains the information that needs to be attended to. The attention weights for each head are multiplied with their corresponding values $V_j$ and concatenated to obtain the output of the multi-head attention:

$$MultiHeadOutput = Concat(AttWeight_1 V_1, ..., AttWeight_h V_h), \qquad (9)$$

the Concat operation concatenates the outputs of different heads. Finally, the output of the multi-head attention is integrated through a linear transformation to obtain the final attention output:

$$Output = MultiHeadOutput \times W^O, \qquad (10)$$

here, $W^O$ is a learned weight matrix.

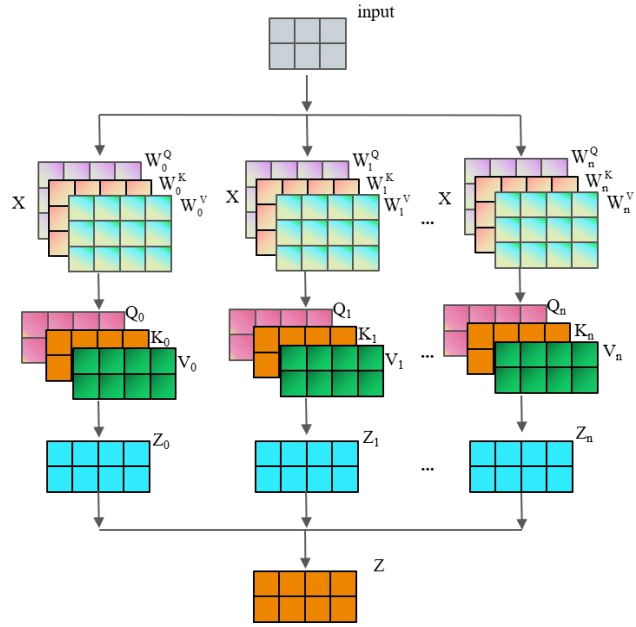

**Figure 3 Multi-head attention**

**The Hybrid Model**

    The overall structure of the hybrid model based on GANs and transformer, as depicted in Figure 4, is proposed in this paper. As explained earlier, GANs play a pivotal role in addressing the challenges of data imbalance and insufficient sample size in the input model. On the other hand, the transformer model is adept at effectively capturing global contextual dependencies, facilitating a comprehensive understanding of the semantic meaning of the input sequence. The uniqueness of this hybrid model lies in its adept

utilization of GANs' adversarial characteristics, continually training and optimizing to generate reliable new samples, thereby enriching sample diversity and increasing data volume. Consequently, the trained model exhibits enhanced performance, better suited to adapt to the complex and ever-changing data distributions encountered in the real world. Furthermore, by incorporating the features of the transformer model, the hybrid approach mitigates potential issues such as vanishing or exploding gradients that commonly afflict traditional sequential models. This characteristic enhances the model's robustness during training and empowers it to excel in handling long-range dependencies.

In summary, the proposed hybrid model skillfully combines the strengths of GANs and transformer, effectively addressing challenges related to data imbalance and insufficient samples in the dataset. Additionally, the model demonstrates outstanding performance in comprehending the semantic information of input sequences and managing global dependencies. Such a fusion makes it a powerful tool with immense potential for various real-world applications.

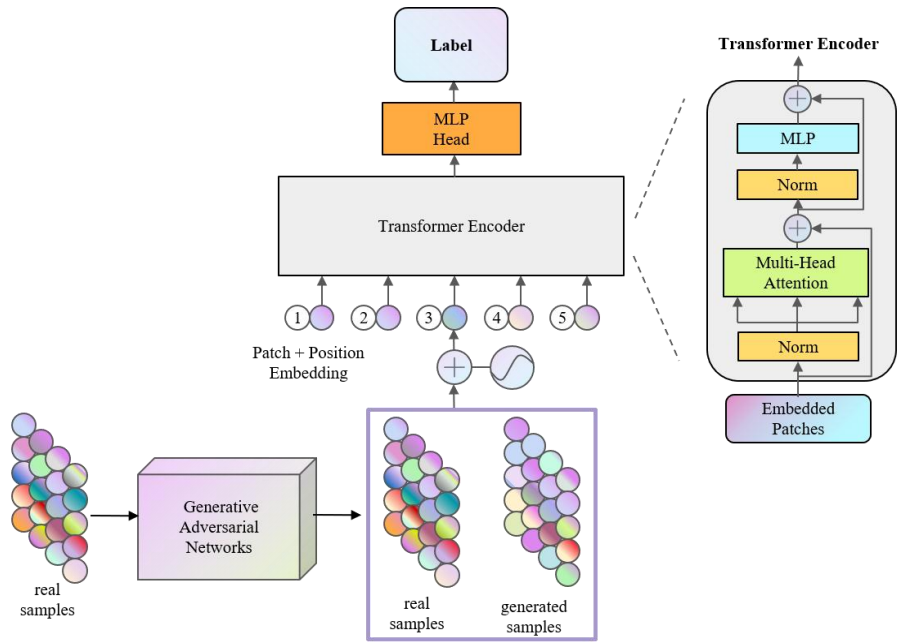

**Figure 4 The framework of the proposed model**

## EXPERIMENTS

### Datasets Description

The proposed hybrid model is evaluated on four datasets typically used in traffic incident detection. In these four datasets, a total of 1600 incident samples and 7240 non-incident samples were selected for the experiment. To address the issue of sample imbalance, Generative Adversarial Networks (GANs) were employed to generate data based on the 1600 incident samples.

- *PeMS Dataset*

    PeMS is a traffic flow database specifically designed for the state of California. It collects real-time data from over 39,000 independent detectors deployed on California highways. This comprehensive dataset includes traffic flow parameters, accident data, and weather information. The PeMS system also provides software components for analyzing and processing the collected data.

- *I-880 Dataset*

    The experimental data came from the famous I880 expressway traffic incident database in the United States. The I880 highway is 9.2 miles long and has three to five lanes, including some with high occupancy. In the northbound lane and southbound lane, there are 18 and 17 loop detection stations respectively. The database completely records the traffic flow speed and occupancy data collected by the loop during the sampling period, as well as the incident location, start and end time, and other information. The database is widely used to validate and evaluate AID algorithms.

- *Whitemud Drive Dataset*

    Whitemud Drive is a 28-kilometer highway in Edmonton, Alberta, Canada, with a speed limit of 80 kilometers per hour. It is equipped with loop detectors on main lanes and ramps to collect traffic parameters such as vehicle speed, flow rate, and occupancy every 20 seconds. This database facilitates traffic flow prediction, model analysis, and research on highway capacity. These information is valuable for urban traffic planning, management, and road safety. The data also serve as references for developing and improving traffic flow models to enhance transportation system efficiency and sustainability.

- *NGSIM Dataset*

    The NGSIM dataset, initiated by the US Federal Highway Administration, collects real-world vehicle trajectory data for research purposes in driver behavior analysis, traffic flow analysis, microsimulation modeling, and vehicle trajectory prediction. Collected using video cameras and radar sensors, the data provides real-time vehicle positions, speeds, accelerations, and more, sampled every 20 seconds for comprehensive traffic flow coverage.

**Evaluation Metrics**

The model is evaluated in five aspects: detection rate, false positive rate, classification rate, and area under the ROC curve (*18*). The formula is as follows:

$$DR = \frac{number\ of\ incident\ samples\ correctly\ detected}{number\ of\ incident\ samples}, \quad (11)$$

$$FAR = \frac{number\ of\ incident\ samples\ falsely\ detected}{number\ of\ the\ incident\ samples\ correctly\ detected}, \quad (12)$$

$$CR = \frac{number\ of\ incident\ samples\ correctly\ detected}{number\ of\ the\ samples}. \quad (13)$$

AUC (Area Under the Curve) is a metric used to evaluate the performance of classifiers and is commonly employed for assessing the predictive accuracy of binary classification models, such as binary classifiers. AUC represents the area under the ROC curve, where ROC stands for Receiver Operating Characteristic. In the ROC curve, different classification thresholds are used to calculate the values of True Positive Rate (TPR), also known as Recall, and False Positive Rate (FPR). TPR denotes the proportion of positive samples correctly classified as positive, while FPR represents the proportion of negative samples incorrectly classified as positive. The ROC curve plots TPR on the vertical axis and FPR on the horizontal axis, thereby displaying the classifier's performance at various thresholds (*19*).

**Synthetic Data**

Using distribution comparison as a means to evaluate the performance of GANs models is a widely adopted and effective approach (*20*). As the objective of GANs models is to generate data samples that closely resemble the distribution of real data, comparing the distributions of generated and real data provides insights into the quality of the generated data and the performance of the model.

Distribution comparison methods provide visualization tools and statistical metrics to compare the distributions of generated and real data (*21*). These methods help us understand the similarity or dissimilarity between the distributions of generated and real data. The Cumulative Distribution Function (CDF) is a function used to describe the probability distribution of a random variable. It calculates the probability that the variable is less than or equal to a given value. Plotting the cumulative distribution function allows us to visually compare the probability distributions of random variables, including measures like mean, variance, skewness, and kurtosis. **Figure 5** depicts the comparison of CDFs between the real data and the generated data for four datasets.

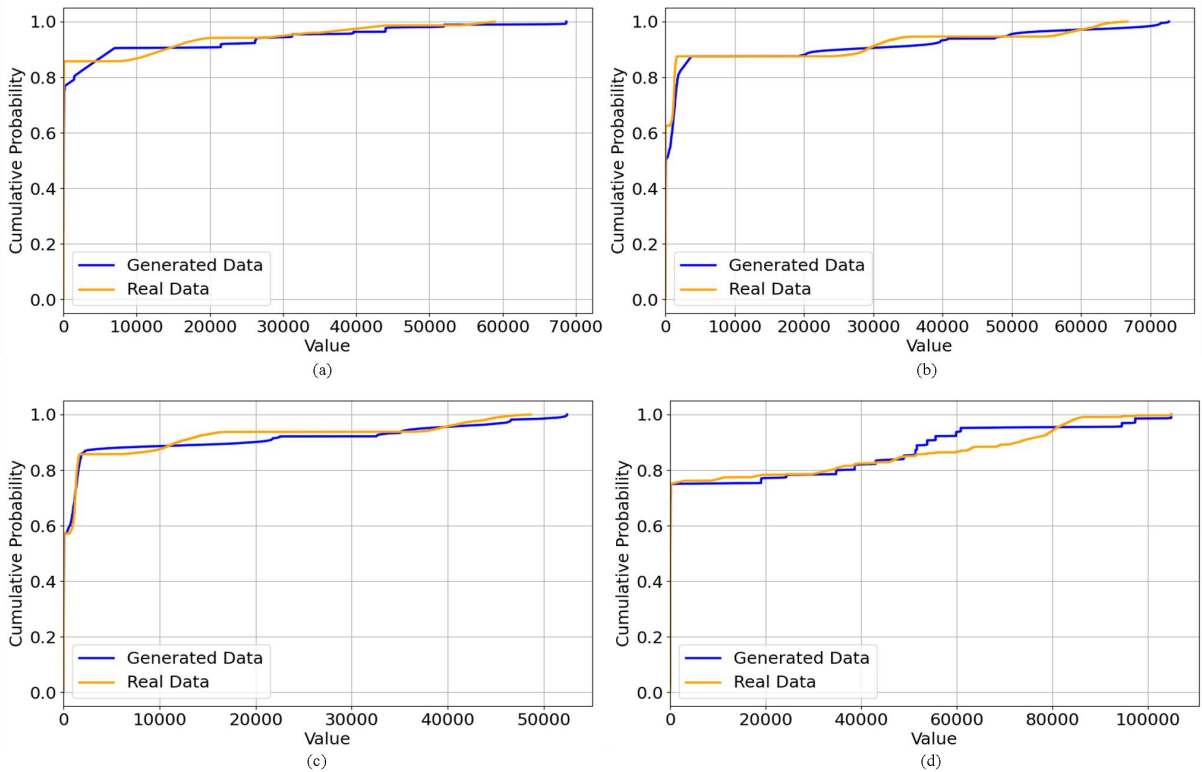

**Figure 5 The comparison of the CDF metric between generated data and real data on different datasets. (a) compares on the PeMS dataset, (b) compares on the I-880 dataset, (c) compares on the Whitemud Drive dataset, and (d) compares on NGSIM dataset**

Kernel Density Estimation (KDE) is a non-parametric technique used to estimate the probability density function of a continuous random variable based on a finite set of observations (*22*). It provides a

*Lu, Zhang, and Xiao*

smooth and continuous estimate of the underlying probability distribution. **Figure 6** compares the KDE distributions between the real data and the generated data from four different datasets.

The standard deviation(SD) is a statistical measure used to assess the dispersion or variability of data points within a dataset. It quantifies the average deviation of data points from the mean of the dataset. **Table 1** presents the medians, means, and standard deviation of four datasets along with the synthetic dataset generated from them.

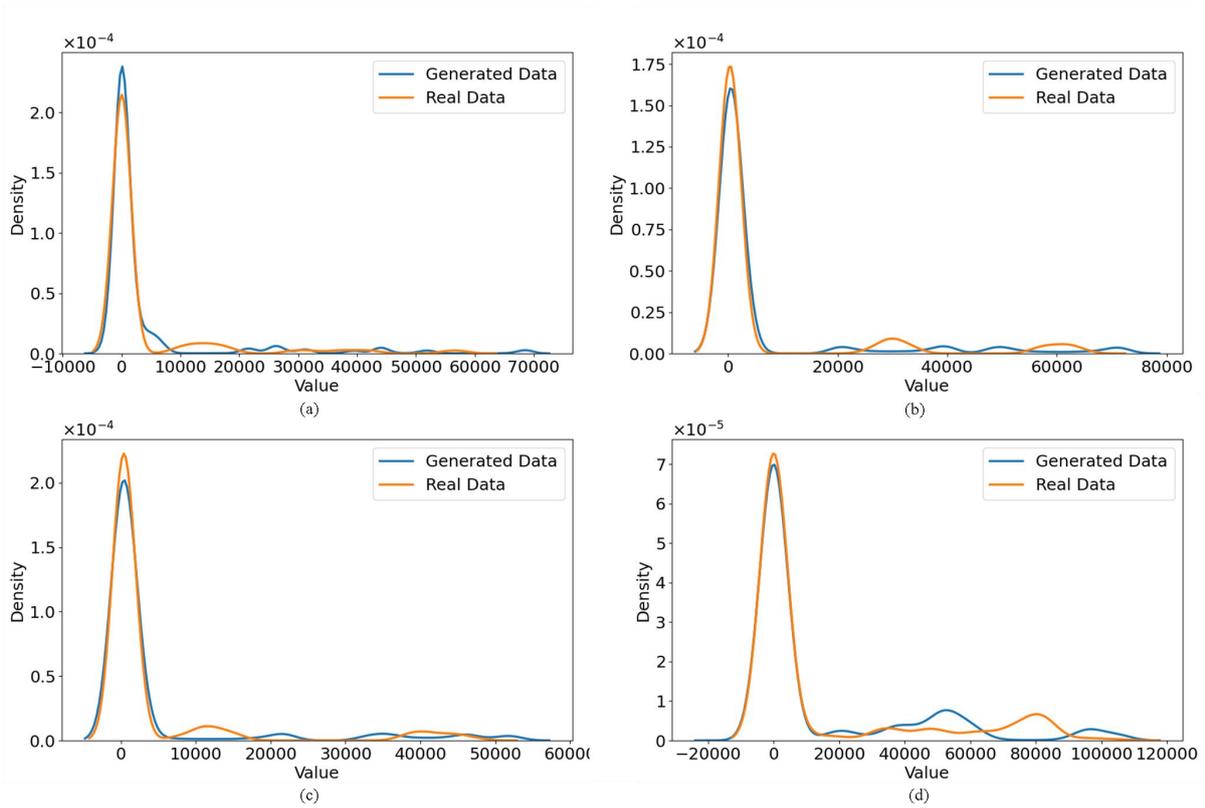

**Figure 6** The comparison of the KDE metric between generated data and real data on different datasets. (a) compares on PeMS dataset, (b) compares on I-880 dataset, (c) compares on Whitemud Drive dataset, and (d) compares on NGSIM dataset

**Table 1 Comparison of Distribution between Generated Data and Real Data**

| Dataset | PeMS | | | I-880 | | | Whitemud Drive | | | NGSIM | | |
|---|---|---|---|---|---|---|---|---|---|---|---|---|
| | medians | means | SD | medians | means | SD | medians | means | SD | medians | means | SD |
| **Real** | 13.6 | 3570.9 | 10332.6 | 52 | 6632.2 | 16608.1 | 50.6 | 4456.0 | 11720.2 | 35.5 | 14230 | 27847.3 |
| **Synthetic** | 14.1 | 3575.4 | 10512.5 | 53.7 | 6750.2 | 16784.7 | 52.1 | 4238.2 | 11240.0 | 39.9 | 13960 | 26526.7 |

These results provide further evidence of the GANs model's effectiveness in capturing and reflecting the correlations between variables in the raw samples. By training the GANs model on the raw data, it learns the underlying patterns and dependencies present in the data distribution. As a result, when generating new samples, the GANs can generate data points that not only exhibit similar individual variable distributions but also preserve the inter-variable relationships observed in the original samples.

**Performance Comparison between Transformer Model and Baseline Models**

*Lu, Zhang, and Xiao*

In this section, the real datasets are employed in the experiments to demonstrate the superior performance of the transformer model in the domain of binary classification. Considering the realistic scenario, most datasets suffer from severe class imbalance. Therefore, for the comparative experiments, 7,240 real non-incident samples and 1,600 real incident samples are selected from the aforementioned four datasets to validate the superiority of the transformer model. All the dataset was split into a training set and a test set, with a ratio of 6:4. The comparative models include the BPNN model, CNN model, CT model, LSTM model, RNN model, WaveNet model, and TCN model. The evaluation metrics used include detection rate, false alarm rate, F1 score, and classification rate. The BPNN model used in this study is a three-layer backpropagation neural network with a specific structure of n x 4n x 1. This means that the network's input layer has n nodes, the hidden layer has 4n nodes, and **th**e output layer has 1 node. Here, "n" refers to the number of input variables, which in the case of the PeMS dataset is 7. Additionally, in the ensemble learning of WaveNet and TCN, the weight parameter "N" is set to 0.5. The performances of the transformer models and the previously commonly used models trained on 4 datasets are listed in **Table 2** and **Table 3**. **Table 2** shows the performance comparison between the transformer model and the baseline models on the PeMS dataset and the I-880 dataset. **Table 3** shows the performance comparison on the Whitemud Drive dataset and the NGSIM dataset. It is important to note that the bolded values represent the highest values achieved for each evaluation metric. The results clearly demonstrate the superior performance of the transformer model in terms of detection rate and False Alarm Rate in the task of traffic incident detection.

**Table 2 Performance Comparison on PeMS Dataset and I-880 Dataset**

| Dataset | PeMS | | | | | I-880 | | | | |
|---|---|---|---|---|---|---|---|---|---|---|
| Models | DR | FAR | CR | AUC | ADT(s) | DR | FAR | CR | AUC | ADT(s) |
| BPNN | 0.6123 | 0.0842 | 0.9136 | 0.8558 | 31 | 0.91 | 0.02 | 0.97 | 0.982 | 53 |
| CNN | 0.8647 | 0.0895 | 0.8872 | 0.9024 | 24 | 0.92185 | 0.01788 | 0.91964 | 0.9714 | 26 |
| CT | 0.8381 | 0.0729 | 0.8581 | 0.882 | **21** | 0.923 | 0.013 | 0.937 | 0.964 | **11** |
| LSTM | 0.9034 | 0.1002 | 0.9024 | 0.9005 | 27 | 0.9665 | 0.02063 | 0.937 | 0.971 | 27 |
| RNN | 0.8776 | 0.1022 | 0.8882 | 0.8942 | 25 | 0.956 | 0.01994 | 0.938 | 0.976 | 22 |
| Wave-net | 0.8567 | 0.1623 | **0.9143** | 0.8411 | 22 | 0.99362 | 0.01031 | 0.973437 | 0.9904 | 23 |
| TCN | 0.8856 | 0.0618 | 0.9127 | 0.9283 | 139 | 0.9941 | 0.006875 | 0.983 | **0.9915** | 137 |
| Transformer | **0.9243** | **0.0566** | 0.8740 | **0.9369** | 24 | **0.999** | **0.002** | **0.999** | 0.966 | 17 |

**Table 3 Performance Comparison on Whitemud Drive Dataset and NGSIM Dataset**

| Dataset | Whitemud Drive | | | | | NGSIM | | | | |
|---|---|---|---|---|---|---|---|---|---|---|
| Models | DR | FAR | CR | AUC | ADT(s) | DR | FAR | CR | AUC | ADT(s) |
| BPNN | 0.892 | 0.013 | 0.912 | 0.902 | 32 | 0.907 | 0.018 | 0.926 | 0.954 | 29 |
| CNN | 0.92344 | 0.02303 | 0.9096 | 0.9675 | 27 | 0.97926 | 0.02819 | 0.976 | 0.9731 | 24 |
| CT | 0.899 | 0.016 | 0.920 | 0.911 | 24 | 0.921 | 0.029 | 0.911 | 0.901 | 23 |
| LSTM | 0.8995 | 0.01639 | 0.943 | 0.9519 | 24 | 0.9808 | 0.02647 | 0.977 | 0.9748 | 25 |
| RNN | 0.9505 | 0.03575 | 0.957 | 0.9618 | 23 | 0.9824 | 0.0275 | 0.977 | 0.9743 | 24 |
| Wave-net | 0.9409 | 0.02372 | 0.959 | 0.97 | 27 | 0.96650 | 0.02303 | 0.972 | 0.9751 | **22** |
| TCN | 0.92982 | 0.006188 | **0.962** | 0.9825 | 141 | 0.9808 | 0.02647 | **0.978** | 0.9748 | 25 |
| Transformer | **0.9730** | **0.0059** | 0.9448 | **0.9955** | **21** | **0.9830** | **0.0012** | 0.9531 | **0.9881** | 22 |

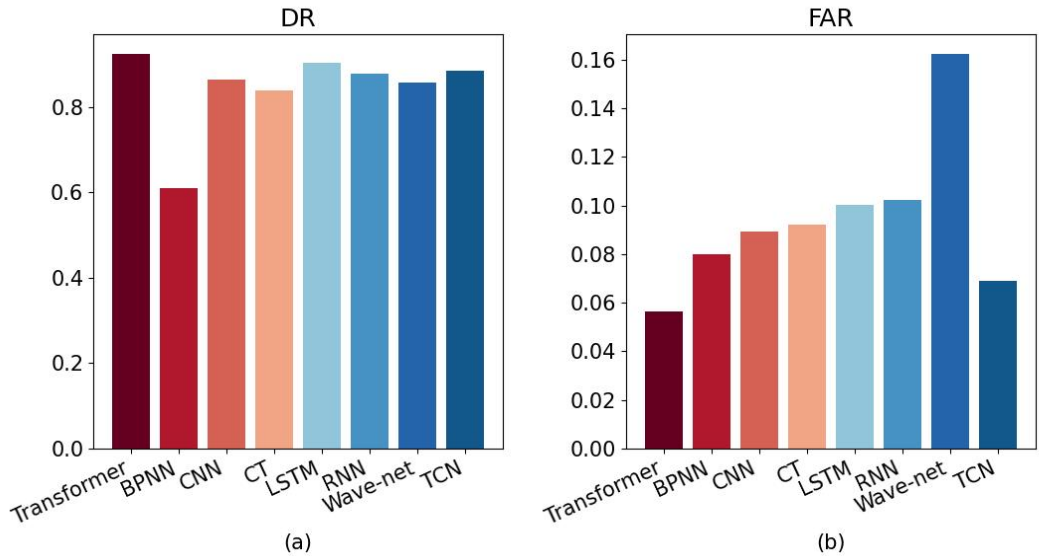

**Figure 7** Performance comparison between transformer model and baseline models: (a) shows the comparison of DR; (b) shows the comparison of FAR

Based on the experimental results, it can be observed that the transformer model consistently outperforms other models and achieves superior performance across all evaluation metrics. Particularly, the transformer model demonstrates the highest performance in terms of detection rate and false alarm rate, which are crucial in traffic incident detection. Additionally, the TCN model also exhibits excellent performance; however, it significantly lags behind other models in terms of average detection time. In real-time traffic incident detection, timeliness is of paramount importance. Therefore, the transformer model emerges as the most optimal and comprehensive choice. **Figure 7** illustrates the bar chart comparing the DR and FAR metrics of different models on the PeMS dataset, clearly demonstrating the superior performance of the transformer model.

**Performance Comparison between Proposed Model and Baseline Models**

In this section, the proposed hybrid model combines the GANs model with the transformer model, where the GANs model is responsible for augmenting the imbalanced dataset into a balanced one. To validate the importance of a balanced dataset in traffic incident detection models, the GANs model is used to augment the four datasets into different incident-to-non-incident ratios, specifically 1:4, 2:3, and 1:1. The augmented datasets are then used for comparative experiments with the same models as in the previous section. This approach aims to demonstrate the significance of a balanced dataset in the context of traffic incident detection. It can be seen from the results that the model trained on the balanced data set gets the better performance, especially in the false positive rate, and gets the ideal reduction. Reducing the false positives rate has always been an important concern of the dichotomy problem. It can be observed that a higher detection rate will inevitably lead to a higher false positive rate. When evaluating a new model, these four metrics need to be balanced for debugging. However, the new model proposed in this paper has good performance in every aspect. Moreover, the new model can transform an unbalanced data set into a balanced data set, which solves an important concern of binary problems. The comprehensive detection performance of the new model is better than that of the baseline models in all aspects.

**Table 4 Performance Comparison of Transformer and Baseline Models on PeMS Dataset**

| Dataset | Models | DR<br>1:4/2:3/1:1 | FAR<br>1:4/2:3/1:1 | CR<br>1:4/2:3/1:1 | AUC<br>1:4/2:3/1:1 | ADT(s)<br>1:4/2:3/1:1 |
|---|---|---|---|---|---|---|
| PeMS | BPNN | 0.753/0.812/0.911 | 0.196/ 0.178/0.11 | 0.833/0.921/0.933 | 0.821/0.918/0.924 | 29/28/26 |
| | CNN | 0.772/0.929/0.936 | 0.122/0.114/0.098 | 0.857/0.904/0.914 | 0.825/0.908/0.914 | 29/24/24 |
| | CT | 0.712/0.733/0.779 | 0.184/0.161/0.121 | 0.783/0.811/0.834 | 0.765/0.835/0.867 | 29/29/26 |
| | LSTM | 0.808/0.926/0.937 | 0.126/0.115/0.10 | 0.861/0.902/0.915 | 0.841/0.906/0.915 | 27/ 25/24 |
| | RNN | 0.701/0.929/ 0.947 | 0.111/0.110/0.097 | 0.852/0.906/0.909 | 0.795/0.909/0.908 | 24/23/**21** |
| | Wave-net | 0.848/0.908/0.939 | 0.065/0.057/0.052 | 0.913/0.929/0.939 | 0.891/0.926/0.939 | 28/22/24 |
| | TCN | 0.849/0.9341/0.960 | 0.060/0.044/0.049 | 0.922/0.947/0.956 | 0.894/0.945/0.956 | 140/142/132 |
| | Transformer | 0.921/0.967/**0.985** | 0.039/0.032/**0.021** | 0.918/**0.966**/0.966 | 0.895/0.992/**0.999** | 23/22/**21** |

**Table 5 Performance Comparison of Transformer and Baseline Models on I-880 Dataset**

| Dataset | Models | DR<br>1:4/2:3/1:1 | FAR<br>1:4/2:3/1:1 | CR<br>1:4/2:3/1:1 | AUC<br>1:4/2:3/1:1 | ADT(s)<br>1:4/2:3/1:1 |
|---|---|---|---|---|---|---|
| I-880 | BPNN | 0.92/0.931/0.97 | 0.018/0.019/0.02 | 0.981/0.981/0.983 | 0.97/0.971/0.98 | 37/39/39 |
| | CNN | 0.959/0.941/0.943 | 0.042/0.039/0.031 | 0.951/0.953/0.956 | 0.948/0.951/0.957 | 26/24/24 |
| | CT | 0.926/0.931/0.938 | 0.026/0.024/0.022 | 0.934/0.939/0.949 | 0.9372/0.941/0.944 | 21/19/19 |
| | LSTM | 0.963/0.937/0.952 | 0.021/0.022/0.033 | 0.976/0.961/0.982 | 0.939/ 0.957/0.960 | 28/27/28 |
| | RNN | 0.973/0.914/0.933 | 0.028/0.016/0.019 | 0.972/0.955/0.956 | 0.932/ 0.949/0.957 | 24/22/21 |
| | Wave-net | 0.982/0.968/0.973 | 0.014/0.019/0.021 | 0.985/0.976/0.978 | 0.984/0.974/0.977 | 24/23/23 |
| | TCN | 0.992/0.976/0.978 | 0.0069/0.009/0.006 | 0.994/0.986/0.989 | 0.983/0.984/**0.990** | 136/135/147 |
| | Transformer | 0.993/ 0.998/**0.999** | 0.002/0.002/**0.0017** | 0.996/**0.996**/0.981 | 0.977/0.981/**0.986** | 21/20/**16** |

**Table 6 Performance Comparison of Transformer and Baseline Models on Whitemud Drive Dataset**

| Dataset | Models | DR<br>1:4/2:3/1:1 | FAR<br>1:4/2:3/1:1 | CR<br>1:4/2:3/1:1 | AUC<br>1:4/2:3/1:1 | ADT(s)<br>1:4/2:3/1:1 |
|---|---|---|---|---|---|---|
| Whitemud Drive | BPNN | 0.893/0.911/0.921 | 0.021/0.017/0.016 | 0.932/0.948/0.950 | 0.931/0.941/0.952 | 31/29/27 |
| | CNN | 0.908/0.911/0.922 | 0.016/0.019/0.018 | 0.969/0.969/0.951 | 0.946/0.946/0.951 | 24/24/22 |
| | CT | 0.872/0.889/0.916 | 0.023/0.019/0.017 | 0.909/0.911/0.913 | 0.903/0.911/0.921 | 26/24/24 |
| | LSTM | 0.935/0.961/0.954 | 0.032/0.032/0.026 | 0.962/0.964/0.964 | 0.952/0.962/0.964 | 27/25/28 |
| | RNN | 0.953/0.948/0.938 | 0.039/0.027/0.024 | 0.958/0.962/0.966 | 0.957/0.960/0.967 | 22/23/20 |
| | Wave-net | 0.946/0.943/0.955 | 0.026/0.018/0.016 | 0.968/0.969/0.961 | 0.960/0.962/0.962 | 24/24/24 |
| | TCN | 0.973/0.975/0.981 | 0.014/0.009/0.008 | 0.987/0.981/0.978 | 0.982/0.980/0.977 | 140/136/137 |
| | Transformer | 0.991/0.993/**0.999** | 0.005/0.003/**0.002** | 0.996/0.996/**0.998** | 0.963/0.997/**0.999** | 20/**19**/**19** |

Table 7 Performance Comparison of Transformer and Baseline Models on NGSIM Dataset

| Dataset | Models | DR 1:4/2:3/1:1 | FAR 1:4/2:3/1:1 | CR 1:4/2:3/1:1 | AUC 1:4/2:3/1:1 | ADT(s) 1:4/2:3/1:1 |
|---|---|---|---|---|---|---|
| NGSIM | BPNN | 0.899/0.913/0.931 | 0.022/0.019/0.013 | 0.912/0.942/0.951 | 0.911/0.923/0.932 | 31/29/27 |
| | CNN | 0.971/0.982/0.985 | 0.017/0.016/0.015 | 0.981/0.972/0.975 | 0.974/0.978/0.979 | 25/22/22 |
| | CT | 0.892/0.899/0.926 | 0.023/0.016/0.013 | 0.919/0.931/0.937 | 0.908/0.911/0.931 | 26/24/24 |
| | LSTM | 0.9770.978//0.981 | 0.026/0.025/0.022 | 0.975/0.979/0.981 | 0.971/0．974/0.983 | 25/25/24 |
| | RNN | 0.967/0.978/0.987 | 0.016/0.016//0.015 | 0.971/0.974/0.976 | 0.976/0.978/0.979 | 25/22/24 |
| | Wave-net | 0.977/0.981/0.985 | 0.024/0.023 | 0.976/0.967/0.975 | 0.977/0.978/0.979 | 24/23/22 |
| | TCN | 0.983/0.982/0.986 | 0.021/0.020/0.019 | 0.979/0.981/0.983 | 0.973/0.981/0.983 | 140/149/143 |
| | Transformer | 0.986/0.987/**0.988** | 0.0012/0.0011/**0.001** | 0.994/**0.996/0.996** | 0.985/0.989/**0.991** | 23/22/**21** |

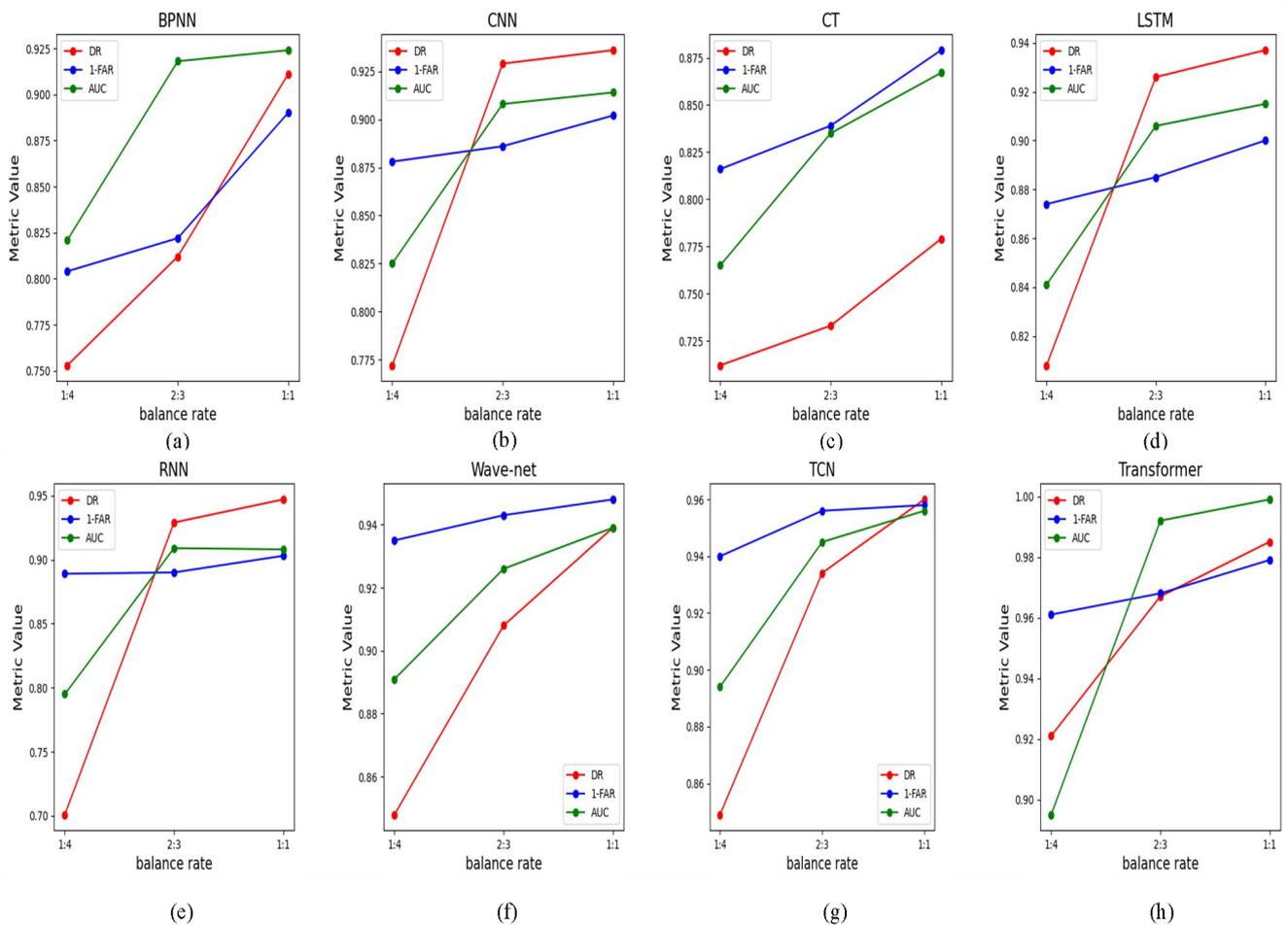

**Figure 8** The comparison of DR, FAR, and AUC over different balance rates on the PeMS dataset between the proposed model and baseline models

**Tables 4, 5, 6, and 7** respectively present the performance metrics of the proposed model and baseline models on four datasets at balance ratios of 1:4, 2:3, and 1:1. The best results are highlighted in bold. From the results, it is evident that the transformer model performs the best comprehensively at a balanced ratio of 1:1 on all four datasets, especially excelling in reducing the FAR. Figure 8 shows the line chart of DR, FAR, and AUC metrics for each model on the PeMS dataset at balance ratios of 1:4, 2:3, and 1:1. It can be clearly seen from **Figure 8** and the tables that as the dataset approaches to balance, the performance of each model improves. Moreover, on balanced datasets, all models exhibit significantly better performance. Furthermore, since the transformer model outperforms the baseline models by a wide margin, the proposed hybrid model combining transformer and GANs exhibits considerable advantages and holds great promise for future applications.

**CONCLUSION**

Traffic incident detection plays a vital role in the realm of intelligent transportation systems by enabling prompt response planning and prevention of secondary accidents. However, challenges such as limited dataset richness and difficulties in collecting real-world incident data have persisted in the field. This study approaches traffic incident detection as a binary classification problem and introduces a method to address issues of data imbalance and scarcity by utilizing the transformer model, a cutting-edge deep learning approach, for traffic incident monitoring. Furthermore, a hybrid model that combines Generative Adversarial Networks and the transformer model is proposed. The evaluation of the model is conducted on four datasets: PeMS, I-880, Whitemud Drive, and NGSIM. The results demonstrate improved detection accuracy and reduced false positive rates. GANs are employed to transform the dataset into a balanced representation, while the transformer model is leveraged to extract relevant features and facilitate traffic incident detection. The experimental findings indicate the efficacy of GANs in generating diverse and augmented samples, effectively addressing the challenges of limited samples and data imbalance. Additionally, the transformer model captures temporal and spatial correlations in the traffic data effectively.

Future research endeavors may explore the diversified variable selection, including holidays and adverse weather conditions, to enhance the model's generalizability. Moreover, the model's application to datasets from other urban systems would contribute to obtaining broader insights. While this study primarily utilizes highway datasets characterized by straight road segments, the inclusion of data from urban road networks is desired in future investigations.

**ACKNOWLEDGMENTS**

The authors would like to express their most sincere gratitude to the China NSFC Program under Grant NO. (61603257, 61906121) who have contributed to the successful completion of this research.

**AUTHOR CONTRIBUTIONS**
The authors confirm contribution to the paper as follows: study conception and design: L. X., X. J.; data collection: L. X., X. J.; analysis and interpretation of results: L. X., Z. D.; draft manuscript preparation: L. X., Z. D. All authors reviewed the results and approved the final version of the manuscript.